\documentclass[10pt,twocolumn,letterpaper]{article}

\usepackage[pagenumbers]{cvpr} 
\usepackage[accsupp]{axessibility}
\usepackage{booktabs}
\usepackage{xcolor}
\usepackage{colortbl}
\usepackage{bbding}
\usepackage{subcaption} 

\usepackage{subcaption}

\definecolor{cvprblue}{rgb}{0.21,0.49,0.74}
\usepackage[pagebackref,breaklinks,colorlinks,allcolors=cvprblue]{hyperref}

\def\paperID{39853} 
\def\confName{CVPR}
\def\confYear{2026}

\title{Detecting Precise Hand Touch Moments in Egocentric Video}

\author{Huy Anh Nguyen \quad Feras Dayoub \quad Minh Hoai\\
Australian Institute for Machine Learning, Adelaide University, Australia
\\ {\tt\small \{huyanh.nguyen, feras.dayoub, minhhoai.nguyen\}@adelaide.edu.au}
}

\begin{document}
\maketitle
\def\mA{\mathcal{A}}
\def\mB{\mathcal{B}}
\def\mC{\mathcal{C}}
\def\mD{\mathcal{D}}
\def\mE{\mathcal{E}}
\def\mF{\mathcal{F}}
\def\mG{\mathcal{G}}
\def\mH{\mathcal{H}}
\def\mI{\mathcal{I}}
\def\mJ{\mathcal{J}}
\def\mK{\mathcal{K}}
\def\mL{\mathcal{L}}
\def\mM{\mathcal{M}}
\def\mN{\mathcal{N}}
\def\mO{\mathcal{O}}
\def\mP{\mathcal{P}}
\def\mQ{\mathcal{Q}}
\def\mR{\mathcal{R}}
\def\mS{\mathcal{S}}
\def\mT{\mathcal{T}}
\def\mU{\mathcal{U}}
\def\mV{\mathcal{V}}
\def\mW{\mathcal{W}}
\def\mX{\mathcal{X}}
\def\mY{\mathcal{Y}}
\def\mZ{\mathcal{Z}}

\def\bbN{\mathbb{N}}
\def\bbR{\mathbb{R}}
\def\bbP{\mathbb{P}}
\def\bbQ{\mathbb{Q}}
\def\bbE{\mathbb{E}}

\def\1n{\mathbf{1}_n}
\def\0{\mathbf{0}}
\def\1{\mathbf{1}}

\def\A{{\bf A}}
\def\B{{\bf B}}
\def\C{{\bf C}}
\def\D{{\bf D}}
\def\E{{\bf E}}
\def\F{{\bf F}}
\def\G{{\bf G}}
\def\H{{\bf H}}
\def\I{{\bf I}}
\def\J{{\bf J}}
\def\K{{\bf K}}
\def\L{{\bf L}}
\def\M{{\bf M}}
\def\N{{\bf N}}
\def\O{{\bf O}}
\def\P{{\bf P}}
\def\Q{{\bf Q}}
\def\R{{\bf R}}
\def\S{{\bf S}}
\def\T{{\bf T}}
\def\U{{\bf U}}
\def\V{{\bf V}}
\def\W{{\bf W}}
\def\X{{\bf X}}
\def\Y{{\bf Y}}
\def\Z{{\bf Z}}

\def\a{{\bf a}}
\def\b{{\bf b}}
\def\c{{\bf c}}
\def\d{{\bf d}}
\def\e{{\bf e}}
\def\f{{\bf f}}
\def\g{{\bf g}}
\def\h{{\bf h}}
\def\i{{\bf i}}
\def\j{{\bf j}}
\def\k{{\bf k}}
\def\l{{\bf l}}
\def\m{{\bf m}}
\def\n{{\bf n}}
\def\o{{\bf o}}
\def\p{{\bf p}}
\def\q{{\bf q}}
\def\r{{\bf r}}
\def\s{{\bf s}}
\def\t{{\bf t}}
\def\u{{\bf u}}
\def\v{{\bf v}}
\def\w{{\bf w}}
\def\x{{\bf x}}
\def\y{{\bf y}}
\def\z{{\bf z}}

\def\balpha{\mbox{\boldmath{$\alpha$}}}
\def\bbeta{\mbox{\boldmath{$\beta$}}}
\def\bdelta{\mbox{\boldmath{$\delta$}}}
\def\bgamma{\mbox{\boldmath{$\gamma$}}}
\def\blambda{\mbox{\boldmath{$\lambda$}}}
\def\bsigma{\mbox{\boldmath{$\sigma$}}}
\def\btheta{\mbox{\boldmath{$\theta$}}}
\def\bomega{\mbox{\boldmath{$\omega$}}}
\def\bxi{\mbox{\boldmath{$\xi$}}}
\def\bnu{\mbox{\boldmath{$\nu$}}}
\def\bphi{\mbox{\boldmath{$\phi$}}}
\def\bmu{\mbox{\boldmath{$\mu$}}}

\def\bDelta{\mbox{\boldmath{$\Delta$}}}
\def\bOmega{\mbox{\boldmath{$\Omega$}}}
\def\bPhi{\mbox{\boldmath{$\Phi$}}}
\def\bLambda{\mbox{\boldmath{$\Lambda$}}}
\def\bSigma{\mbox{\boldmath{$\Sigma$}}}
\def\bGamma{\mbox{\boldmath{$\Gamma$}}}

\newcommand{\myprob}[1]{\mathop{\mathbb{P}}_{#1}}

\newcommand{\myexp}[1]{\mathop{\mathbb{E}}_{#1}}

\newcommand{\mydelta}[1]{1_{#1}}

\newcommand{\myminimum}[1]{\mathop{\textrm{minimum}}_{#1}}
\newcommand{\mymaximum}[1]{\mathop{\textrm{maximum}}_{#1}}
\newcommand{\mymin}[1]{\mathop{\textrm{minimize}}_{#1}}
\newcommand{\mymax}[1]{\mathop{\textrm{maximize}}_{#1}}
\newcommand{\mymins}[1]{\mathop{\textrm{min.}}_{#1}}
\newcommand{\mymaxs}[1]{\mathop{\textrm{max.}}_{#1}}
\newcommand{\myargmin}[1]{\mathop{\textrm{argmin}}_{#1}}
\newcommand{\myargmax}[1]{\mathop{\textrm{argmax}}_{#1}}
\newcommand{\myst}{\textrm{s.t. }}

\newcommand{\denselist}{\itemsep -1pt}
\newcommand{\sparselist}{\itemsep 1pt}

\definecolor{pink}{rgb}{0.9,0.5,0.5}
\definecolor{purple}{rgb}{0.5, 0.4, 0.8}
\definecolor{gray}{rgb}{0.3, 0.3, 0.3}
\definecolor{mygreen}{rgb}{0.2, 0.6, 0.2}

\newcommand{\cyan}[1]{\textcolor{cyan}{#1}}
\newcommand{\blue}[1]{\textcolor{blue}{#1}}
\newcommand{\magenta}[1]{\textcolor{magenta}{#1}}
\newcommand{\pink}[1]{\textcolor{pink}{#1}}
\newcommand{\green}[1]{\textcolor{green}{#1}}
\newcommand{\gray}[1]{\textcolor{gray}{#1}}
\newcommand{\mygreen}[1]{\textcolor{mygreen}{#1}}
\newcommand{\purple}[1]{\textcolor{purple}{#1}}

\definecolor{greena}{rgb}{0.4, 0.5, 0.1}
\newcommand{\greena}[1]{\textcolor{greena}{#1}}

\definecolor{bluea}{rgb}{0, 0.4, 0.6}
\newcommand{\bluea}[1]{\textcolor{bluea}{#1}}
\definecolor{reda}{rgb}{0.6, 0.2, 0.1}
\newcommand{\reda}[1]{\textcolor{reda}{#1}}

\def\changemargin#1#2{\list{}{\rightmargin#2\leftmargin#1}\item[]}
\let\endchangemargin=\endlist

\newcommand{\cm}[1]{}

\newcommand{\mhoai}[1]{{\color{blue}{[MH: #1]}}}
\newcommand{\ha}[1]{{\color{red}{[HA: #1]}}}
\newcommand{\rev}[1]{{\color{mygreen}{#1}}}

\newcommand{\tb}[1]{\textbf{#1}}
\newcommand{\ul}[1]{\underline{#1}}
\newcommand{\gain}[1]{\scriptsize\textcolor{blue}{(#1)}}
\newcommand{\lose}[1]{\scriptsize\textcolor{red}{(#1)}}

\newcommand{\mtodo}[1]{{\color{red}$\blacksquare$\textbf{[TODO: #1]}}}
\newcommand{\myheading}[1]{\vspace{1ex}\noindent \textbf{#1}}
\newcommand{\htimesw}[2]{\mbox{$#1$$\times$$#2$}}


\newif\ifshowsolution
\showsolutiontrue

\ifshowsolution
\newcommand{\Comment}[1]{\paragraph{\bf $\bigstar $ COMMENT:} {\sf #1} \bigskip}
\newcommand{\Solution}[2]{\paragraph{\bf $\bigstar $ SOLUTION:} {\sf #2} }
\newcommand{\Mistake}[2]{\paragraph{\bf $\blacksquare$ COMMON MISTAKE #1:} {\sf #2} \bigskip}
\else
\newcommand{\Solution}[2]{\vspace{#1}}
\fi

\newcommand{\truefalse}{
\begin{enumerate}
	\item True
	\item False
\end{enumerate}
}

\newcommand{\yesno}{
\begin{enumerate}
	\item Yes
	\item No
\end{enumerate}
}

\newcommand{\Sref}[1]{Sec.~\ref{#1}}
\newcommand{\Eref}[1]{Eq.~(\ref{#1})}
\newcommand{\Fref}[1]{Fig.~\ref{#1}}
\newcommand{\Tref}[1]{Table~\ref{#1}}

\newcommand{\parens}[1]{\left(#1\right)}
\newcommand{\braces}[1]{\left\{#1\right\}}
\newcommand{\bracks}[1]{\left[#1\right]}
\newcommand{\modulus}[1]{\left\vert#1\right\vert}
\newcommand{\norm}[1]{\left\Vert#1\right\Vert}
\newcommand{\angular}[1]{\langle#1\rangle}
\newcommand{\lmod}{\left|\!\left|}
\newcommand{\rmod}{\right|\!\right|}

\begin{abstract}
We address the challenging task of detecting the precise moment when hands make contact with objects in egocentric videos. This frame-level detection is crucial for augmented reality, human-computer interaction, assistive technologies, and robot learning applications, where contact onset signals action initiation or completion. Temporally precise detection is particularly challenging due to subtle hand motion variations near contact, frequent occlusions, fine-grained manipulation patterns, and the inherent motion dynamics of first-person perspectives.

To tackle these challenges, we propose a Hand-informed Context Enhanced module (HiCE; pronounced `high-see') that leverages spatiotemporal features from hand regions and their surrounding context through cross-attention mechanisms, learning to identify potential contact patterns. Our approach is further refined with a grasp-aware loss and soft label that emphasizes hand pose patterns and movement dynamics characteristic of touch events, enabling the model to distinguish between near-contact and actual contact frames. We also introduce TouchMoment, an egocentric dataset containing 4,021 videos and 8,456 annotated contact moments spanning over one million frames. Experiments on TouchMoment show that, under a strict evaluation criterion that counts a prediction as correct only if it falls within a two-frame tolerance of the ground-truth moment, our method achieves substantial gains and outperforms state-of-the-art event-spotting baselines by 16.91\% average precision. Code is available at \url{https://github.com/bbvisual/hice}.

\end{abstract}

\section{Introduction}

This paper studies the problem of detecting the precise moment when a hand makes contact with an object in egocentric video. Touch moments mark critical temporal boundaries in manipulation—indicating when an action begins, transitions, or completes—and thus provide rich cues for understanding human behavior, assessing motor capability, analyzing skilled performance, and transferring dexterous demonstrations to robots. Egocentric video, collected from head-mounted or wearable cameras, is a natural modality for this task because it directly captures hand motions and the user's visual perspective during manipulation. Reliable identification of touch moments in such recordings can therefore support a wide range of applications that depend on fine-grained temporal detection.

In this work, we focus on detecting the first touch moment—the single frame at which contact between the hand and an object is first established. Touch moment is treated as an instantaneous event rather than a temporal segment. When a video contains multiple interactions, our goal is to detect each of these first-contact moments. We consider only intentional touch events that contribute to the manipulation process, excluding incidental brushes or accidental collisions. Formally, given an input video sequence, the task is to identify a sparse set of frames marking these transitions from non-contact to contact across the sequence.

\begin{figure}
\centering
    \centering
    \includegraphics[width=1\linewidth]{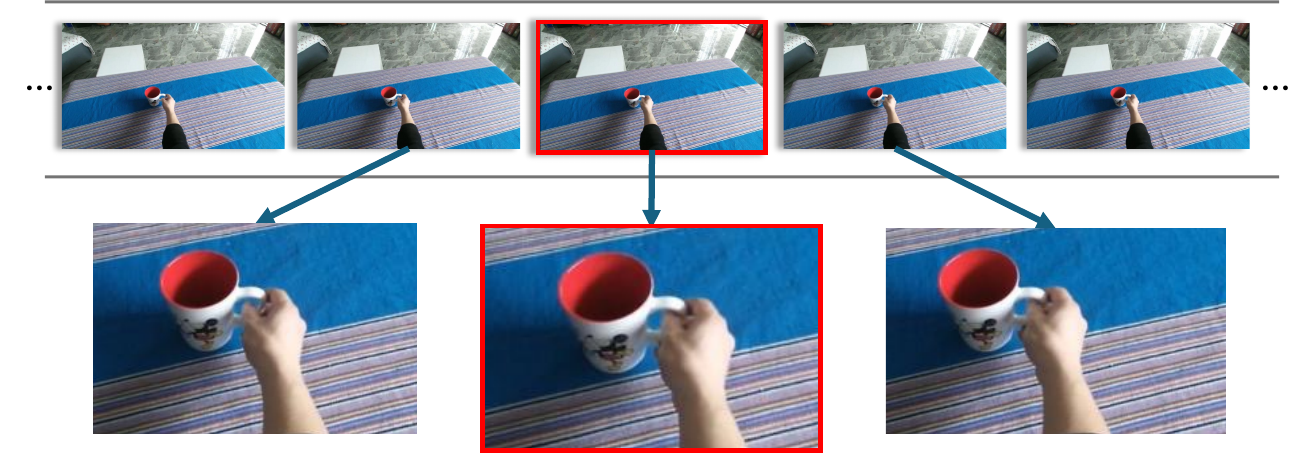}
    \caption{We develop a model to detect the precise touch moment—the exact frame when the hand first makes contact with an object in egocentric video. As illustrated, distinguishing the true touch moment (the frame with a red border) is challenging due to subtle near-touch hand motions, rapid camera movement, and fine-grained grasp changes that make near-touch frames visually similar to the actual moment of contact.
    }
    \label{fig.teaser}
\end{figure}

Detecting the exact touch moment in egocentric video, however, is extremely challenging, as illustrated in \cref{fig.teaser}. Rapid head motion introduces strong egomotion, while approaching hands often create severe occlusions and perspective distortions at close range. Subtle micro-motions immediately before contact, fine-grained pre-touch hand shaping, and motion blur during dynamic activities further obscure the true moment of contact. These factors make the touch moment difficult to distinguish from near-touch frames or other ambiguous transitional states, requiring models that can reliably extract discriminative hand cues despite noise, occlusion, and continuously shifting viewpoints inherent to first-person video.

Precise moment detection is not new in computer vision, and prior work has addressed onset prediction and fine-grained action spotting across several domains. However, these methods are typically developed under assumptions that differ substantially from egocentric hand–object interaction. For example, techniques for detecting the onset of facial action units rely on static or near-static camera setups, making them unsuitable for the strong egomotion intrinsic to first-person video. Likewise, action-spotting methods in sports—such as those used for Olympic events—operate on videos where the target activity dominates the entire frame and the subject remains consistently visible, allowing key moments to be identified without localizing fine-scale regions. In contrast, touch moments in egocentric video occur within compact spatial areas around the fingertips and depend on subtle grasp configurations leading up to contact.

To address these challenges, we introduce the Hand-informed Context Enhancement module (HiCE; pronounced `high-see'), designed to augment frame-level feature extractors with dedicated hand-centric spatiotemporal modeling. Because the cues that signal an impending touch moment are brief and easily diluted in full-frame representations, HiCE explicitly encodes fine-grained hand motion patterns and injects these signals back into the global feature stream to preserve subtle temporal transitions. The module first extracts hand regions using off-the-shelf detectors (or ground truth boxes during training when available) and expands them to include relevant local context. These regions are then processed by a backbone that captures both spatial structure and short-term temporal dynamics from neighboring frames. The resulting hand-specific features are fused with global frame features through cross-attention, guiding the model to attend to contact-critical regions. To further strengthen hand representation learning, we incorporate a grasp loss that leverages grasp predictions from \cite{hands23}.

To support research on frame-accurate touch detection, we also introduce TouchMoment, an egocentric dataset comprising 4,021 videos and 8,456 annotated touch moments across diverse objects, surfaces, and environments. Experiments on TouchMoment show that our HiCE module delivers substantial gains over strong baselines, including under strict evaluation settings where predictions must fall within a two-frame tolerance of the ground-truth touch moment. In short, our contributions are: {\bf (1)} We study and formalize frame-level touch moment detection as a fine-grained egocentric event-spotting task—an important problem for understanding manipulation, skill analysis, and human–robot learning that has received little dedicated attention in prior work. {\bf (2)} We propose HiCE, a hand-centric enhancement module that injects specialized spatiotemporal hand features into event-spotting architectures, leading to substantial improvements in detecting precise touch moments. {\bf (3)} We introduce TouchMoment, a dataset for egocentric touch detection, enabling systematic investigation, comparison, and benchmarking of methods for this emerging task.

\section{Related Works}
\tb{Temporal action localization and event spotting.} Action spotting, introduced by \cite{soccernet, soccernetas}, differs from traditional temporal action localization (TAL) by identifying the precise onset frame of an action rather than predicting start and end boundaries of action segments. While TAL methods such as \cite{sst, gtad, asformer, actionformer, tridet}, typically evaluate using temporal IoU over multi-second intervals, action spotting employs frame-level mAP metrics at tolerance windows ranging from loose (5–60 seconds) to tight (1-5 seconds).
One way to adapt TAL models to spotting is to force dense per-frame predictions through dense anchors or sliding-window proposals. This is inefficient for two reasons. First, achieving frame-level precision requires generating proposals at extremely fine temporal strides, leading to a large number of overlapping windows and substantial computational cost. Second, these proposal mechanisms are designed to regress start and end times of multi-frame temporal segments, which makes them less effective for capturing the subtle, instantaneous transitions characteristic of spotting tasks. As noted in the survey \cite{as_survey}, proposal-based TAL formulations are therefore not ideal for high-precision, frame-level event detection.
As applications demand finer temporal understanding, action spotting has evolved toward tighter tolerances. \cite{e2e_spot} demonstrated that tolerances below 4 frames are necessary for tasks requiring precise temporal boundaries—such as sports performance analysis where millisecond differences matter, skill assessment where action phases must be clearly delineated, and robot learning from human demonstrations where contact timing determines manipulation success. At these sub-second scales, distinguishing the exact moment of ball contact, foot landing, or hand-object touch becomes critical, particularly in egocentric scenarios where rapid, subtle events occur within minimal temporal windows.
Current spotting approaches fall into two categories. Two-phase methods first extract features, then perform temporal localization. \citet{baidu} pioneered this paradigm using an ensemble of five models \cite{tpn, gta, vtn, ircsn, slowfast} trained on SoccerNet, with subsequent works \cite{soares, astra} adopting their pre-extracted features for temporal detection, or using features to distill in \cite{comedian}. While effective for precise spotting, this reliance on fixed features limits adaptability to new domains.
End-to-end methods emerged to address this limitation. E2E-Spot and T-DEED \cite{e2e_spot, tdeed} employ ResNet-Y backbones with temporal gates (GSM\cite{gsm}/GSF\cite{gsf}) and sequence modeling GRU \cite{gru} in E2E-Spot, Scalable-Granularity Perception (SGP-Layer) \cite{tridet} encoder-decoder in T-DEED. Beyond standard classification heads, T-DEED and ASTRA \cite{tdeed, astra} introduced displacement heads that refine predictions along the temporal axis, enabling sub-frame-level localization.
However, these methods process frames uniformly without spatial reasoning about salient regions. UGLF \cite{uglf} addresses this by leveraging vision-language models \cite{glip, glip2} to localize objects and fuse their features via attention \cite{aam}. This approach has two limitations: (1) it requires pre-defining object categories for grounding, and (2) it treats all objects equally. For egocentric touch detection, where contact cues are localized to hand regions, uniform spatial attention is suboptimal. Nevertheless, incorporating spatial object features demonstrates substantial improvements over global frame representations.

\myheading{Hand-object contact analysis.} Hand analysis plays a central role in understanding human behavior and interaction in computer vision. Beyond a large body of work on detecting and tracking the hands themselves (e.g., \cite{m_Narasimhaswamy-etal-ICCV19, m_Huang-etal-CVPR22}), a growing line of research has focused on modeling how hands interact with objects, particularly through contact. A rich set of image-based methods has explored frame-level hand–object contact recognition, showing that identifying whether a hand is in contact provides strong cues for action recognition, affordance reasoning, and grasp understanding. Early works such as 100DOH \cite{100doh}, ContactHands \cite{contacthands_2020}, and more recent methods like Hands23 \cite{hands23} classify contact states or categorize interaction types from single frames, relying primarily on local hand appearance and object context. \citet{sn_bodyhands_cvpr_2022} focus on hand detection and hand–body association, using spatial overlap between hand and body regions to infer hand–body contact, demonstrating how contact cues can be derived indirectly from static imagery.

Beyond predicting binary contact, several works aim to localize where contact occurs on objects. HOT \cite{hot} detects pixel-level human–object contact regions by training a segmentation-based model to predict dense contact heatmaps and body-part labels from a single RGB image, using manually annotated 2D polygons as supervision. This provides a richer spatial description of contact than binary classification but remains a purely appearance-based formulation without temporal reasoning. At a higher level of abstraction, \citet{handprobe} leverage contact maps and grasp types to infer object functionality and affordances, but requires rich supervision and is computationally expensive.

More recently, video-based interaction models such as HOISTFormer \cite{sn_hoist_cvpr_2024} address hand–object detection, segmentation, and tracking in video. While such models can in principle be adapted for touch spotting, they face two challenges: (1) they are trained on either sparsely annotated datasets such as VISOR \cite{ek_visor} or dominated by third-person views, where hand–object motion is more pronounced and contact regions are stable; and (2) in egocentric video, hand motion is subtle, interactions occur at close range, and high-frame-rate videos introduce additional difficulty in distinguishing the exact onset of contact.

While the above approaches highlight the semantic and spatial importance of contact, they operate either on isolated frames or on interaction segments that span many frames. As a result, they lack the temporal granularity needed to discriminate the subtle micro-movements that immediately precede contact, particularly in egocentric video where occlusion, hand jitter, rapid approach motion, and strong egomotion make individual frames highly ambiguous. Consequently, existing contact-understanding methods cannot determine when contact first occurs. This motivates the need for temporally informed, hand-centric models capable of capturing short-term motion cues and distinguishing near-touch states from the precise moment of first contact—the focus of our work.

\section{TouchMoment: An Egocentric Touch Dataset}

Despite the importance of identifying precise touch moments in egocentric video, no existing dataset provides large-scale, frame-level annotations of hand touches suitable for developing and benchmarking methods for this task. To fill this gap, we introduce TouchMoment, a new egocentric video dataset that captures a wide range of everyday manipulation scenarios and supplies exact touch-moment annotations for each interaction. The dataset comprises 4,021 videos sourced from diverse environments, object categories, and interaction contexts, and includes 8,456 manually annotated touch moments recorded at the frame level (\cref{table.data_statistics}). Alongside touch annotations, TouchMoment provides temporal segmentation of interaction sequences and localized hand regions to support hand-centric modeling. The following sections detail the data sources, annotation protocol, and an analysis of the dataset's characteristics.

\subsection{Data Sources}
To build TouchMoment, we source egocentric interaction sequences from two publicly available datasets: HOI4D and the egocentric subset of TACO, both of which contain rich recordings of hand–object manipulation suitable for annotating frame-level touch moments.

HOI4D is a large-scale benchmark for category-level human–object interaction understanding. It provides egocentric RGB-D video sequences captured across diverse indoor environments, featuring a broad range of object categories and interaction types. The dataset includes dense annotations such as 3D hand pose, object pose, panoptic segmentation, and motion segmentation, making it a valuable source of fine-grained hand–object interaction footage.

We further incorporate sequences from the egocentric portion of the TACO dataset, which contains recordings of bimanual tool and object manipulation captured using a head-mounted RGB camera at 30 fps. These videos include continuous interaction patterns such as grasping, re-grasping, and coordinated motion of both hands, offering temporal variety and diverse manipulation contexts that complement the HOI4D clips. Compared to HOI4D, TACO contains longer and more structured interaction sequences, often involving closely coordinated bimanual activity.

From these data sources, we select segments where the hand–object interface is visible, the transition into contact is discernible, and temporal continuity is preserved. These criteria ensure that annotators can reliably identify and label the precise frame corresponding to each touch moment.

\subsection{Touch Annotation}
We define a touch event as the moment when any part of the hand makes physical contact with an object. Annotators determine this moment based on observable changes in hand motion, object motion, or object deformation. We include only intentional and visually unambiguous interactions, and exclude accidental contact or cases where the contact moment cannot be reliably observed. After the moment of touch, the hand remains in contact with the object for a short duration, ensuring the event reflects a meaningful interaction rather than a single-frame alignment.

To ensure consistency, we cross-check annotations and resolve disagreements through direct comparison. For HOI4D, which provides action segment annotations, we manually annotate all touch events in the validation and test splits. For the training split, we manually annotate 10\% of the segments and use these annotations to develop an automatic annotation tool that assigns touch frames to the remaining of the training data. On a held-out subset with full manual annotation, this automatic tool differs from manual labels by an average of 1.94 frames, demonstrating sufficient accuracy for large-scale training data. For TACO, we manually annotate all selected segments. Unlike HOI4D, TACO does not provide action segmentation, and the hand–object contact patterns are more varied in timing and subtlety. As a result, direct manual annotation is required to determine consistent touch boundaries.

\begin{table}[]
\centering
\setlength{\tabcolsep}{4pt}  
\begin{tabular}{@{}lrrrrr@{}}
\toprule
                                        & \multicolumn{3}{c}{\tb{HOI4D}} & \multicolumn{2}{c}{\tb{TACO}} \\
                                        & train   & val    & test   & train       & test       \\ \midrule
\# Frames                                  & 686K  & 41K  & 127K & 156K      & 35K      \\
\# Touch events                            & 4979    & 324    & 930    & 1689        & 364        \\
\# Clips (0 touches)                  & 1       & 0      & 0      & 5           & 0          \\
\# Clips (1 touch)                   & 42      & 0      & 10     & 198         & 89         \\
\# Clips (2 touches)                 & 1937    & 106    & 349    & 738         & 133        \\
\# Clips (3 touches)                 & 180     & 16     & 38     & 5           & 3          \\
\# Clips ($\geq$4 touches) & 128     & 16     & 27     & 0           & 0          \\ \bottomrule
\end{tabular}
\vskip -0.1in
\caption{\textbf{TouchMoment dataset statistics.}}\label{table.data_statistics}
\vskip -0.1in
\end{table}

\section{Proposed Methodology}
In this section, we formalize the frame-level touch detection task and present our approach. We first introduce the problem formulation, then describe the Hand-informed Context Enhancer (HiCE) for augmenting frame features with localized hand cues. We detail the temporal modeling block and prediction heads that operate on these enriched features, and finally present the training supervision used to optimize the full architecture.

\subsection{Problem definition}
We follow the precise event spotting formulation introduced in E2E-Spot \cite{e2e_spot}, temporally precise hand touch event detection takes a sequence of $L$ frames $\mX = \{x_i\}_{i=1}^{L}$  as input and predicts a sparse set of touch events $\{(t, \hat{y}_t)\} \in \mathbb{N} \times \{0,1\}$. A prediction is considered correct if it falls within tolerance $\delta$ frames of the ground truth label and has the correct class.  Since our task falls under Precise Event Spotting (PES), evaluation uses small temporal tolerances. We adopt $\delta \leq 2$, which requires models to localize touch events with near frame-level precision. Our approach builds upon T-DEED \cite{tdeed} by incorporating explicit hand-object context
modeling for egocentric perspective interaction scenarios.

\begin{figure*}
    \centering
    \includegraphics[width=\linewidth]{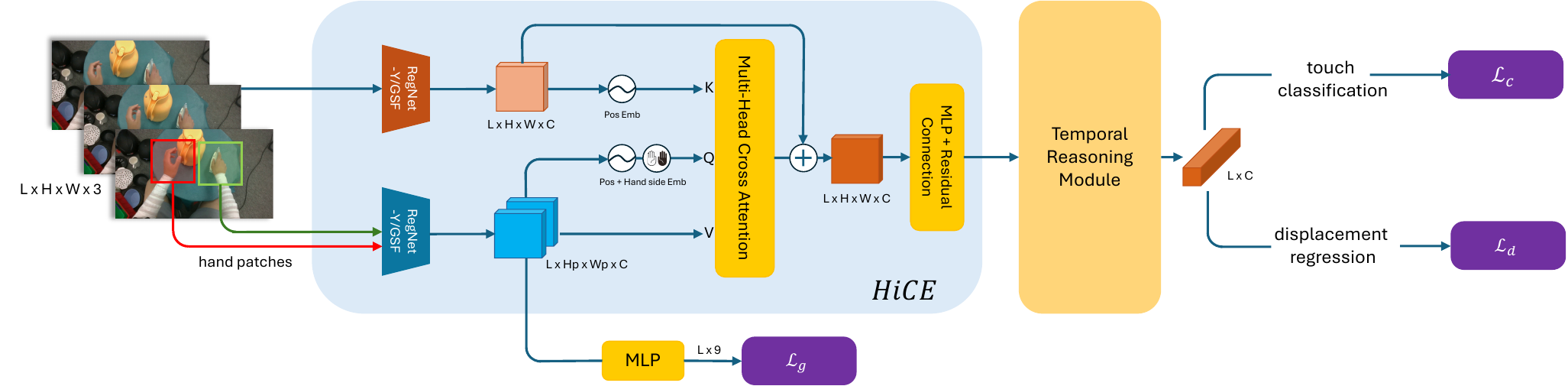}    
    \caption{\tb{The Hand-informed Context Enhanced (HiCE) module} augments the feature extractor with a parallel hand-patch branch that processes left and right hand crops alongside global frame features using RegNet-Y backbones. Hand patches are expanded and encoded with positional and identity embeddings, then used as keys and values in a multi-head cross-attention block where global tokens act as queries. The resulting hand-aware global features are passed into the temporal reasoning module for touch classification and displacement prediction, while the hand features are also used by an auxiliary grasp-prediction head. \label{fig.model}}
\end{figure*}

\subsection{Hand-informed Context Enhanced module}
In standard end-to-end precise action spotting architectures, the model consists of a feature extractor, a temporal reasoning module, and prediction heads, as illustrated in \cref{fig.model}. Our work focuses on strengthening the feature extractor by enriching it with explicit hand–context cues.
To achieve this, we introduce a cross-attention mechanism that allows the global image features to attend selectively to localized hand regions. Given the global feature map
$\mathcal{F} \in \mathbb{R}^{H\times W \times C}$ and the hand patch
features $\mathcal{F}_{\text{lh}}, \mathcal{F}_{\text{rh}} \in
\mathbb{R}^{H_p \times W_p \times C}$, we first augment them with 2D
sinusoidal positional embeddings and hand-identity embeddings (to
distinguish left and right hands), then flatten them into token
sequences. Queries, keys, and values are constructed as
\begin{align}
Q &= f_Q(\mathcal{F} + E_{\text{pos}}), \\
K &= f_K\big([\mathcal{F}_{\text{lh}}, \mathcal{F}_{\text{rh}}]
      + E_{\text{pos}} + E_{\text{id}}\big), \\
V &= f_V([\mathcal{F}_{\text{lh}}, \mathcal{F}_{\text{rh}}]),
\end{align}
where $[\mathcal{F}_{\text{lh}}, \mathcal{F}_{\text{rh}}]$ denotes the
stacked left and right hand features, $E_{\text{pos}}$ are positional
embeddings, $E_{\text{id}}$ are left/right hand identity embeddings, and
$f_Q, f_K, f_V$ are learned linear projections.
We then apply cross-attention in the standard form:
\begin{equation}
\text{CrossAttn}(Q,K,V)
= \text{softmax}\!\left(\frac{QK^{\top}}{\sqrt{d}}\right)V,
\end{equation}
and update the global features via residual connections:
\begin{align}
&\mathcal{F}' = \text{CrossAttn}(Q,K,V) + \mathcal{F}, \\
&\mathcal{F}_{\text{enhanced}} = \text{FFN}(\mathcal{F}') + \mathcal{F}'.
\end{align}
Here $d$ is the feature dimension. As illustrated in \cref{fig.model},
this design allows the global representation to selectively integrate
contact-relevant information from hand regions, producing enhanced
frame-level features enriched with explicit hand context.

\subsection{Temporal Reasoning Module}
For temporal modeling, we adopt the encoder-decoder architecture from T-DEED, which has demonstrated superior performance over sequential approaches like GRU used in E2E-Spot. The T-DEED architecture employs Scalable-Granularity Perception (SGP) layers \cite{tridet} that process features at multiple temporal scales while maintaining high token discriminability—a critical property for precise event localization. The encoder-decoder structure with SGP-Mixer layers enables the model to capture both local and global temporal dependencies while preserving frame-level prediction precision through skip connections that restore the original temporal resolution.

\subsection{Prediction Heads}
We employ a dual-head architecture commonly used in action spotting methods \cite{comedian, soares, astra, tdeed}. The classification head predicts the probability of touch events at each frame using a linear layer followed by softmax activation, generating predictions $\hat{y}^c \in \mathbb{R}^{L \times 2}$ for binary touch classification. The displacement regression head refines temporal localization by predicting frame-level offsets $\hat{y}^d \in \mathbb{R}^{L \times 1}$ to ground truth event timestamps. This dual-head approach enables more precise event localization compared to the label dilation technique employed in E2E-Spot, as it can detect events within a wider temporal window while maintaining high localization accuracy through learned displacement offsets.

Additionally, we integrate a grasp classification head~\cite{wildhands} to provide auxiliary supervision for the hand patch encoder. This module consists of a four-layer MLP that processes concatenated hand patch features $[\mathcal{F_{\text{lh}}}, \mathcal{F_{\text{rh}}}] \in \mathbb{R}^{2 \times H_p \times W_p \times C}$ and predicts hand grasp categories $\hat{y}^g \in \mathbb{R}^{9}$ based on the Cutkosky taxonomy \cite{hands23}, which divides grasps into prehensile and non-prehensile categories. This auxiliary task encourages the hand encoder to learn more discriminative hand representations that are beneficial for touch event detection.

\subsection{Training Supervision}
Given an input clip of \(L\) frames, the model produces predictions \((\hat{y}^c, \hat{y}^d, \hat{y}^g)\), where \(\hat{y}^c \in \mathbb{R}^{L \times 2}\) denotes the classification probabilities for touch events, \(\hat{y}^d \in \mathbb{R}\) represents the regressed displacement offset, and \(\hat{y}^g \in \mathbb{R}^{9}\) corresponds to the 9-class grasp predictions for both hands. Pseudo-labels for grasp classification are obtained from \cite{hands23}. When one or both hands are absent in a frame, the corresponding grasp loss is masked to zero to prevent penalizing missing annotations.

The model is trained with a multi-task objective that jointly optimizes frame-level classification, displacement regression, and grasp classification. The overall training loss is defined as:

{\small
\begin{align}
\mathcal{L}
&= \mathcal{L}_c + \mathcal{L}_d + \lambda_g \mathcal{L}_g \\
&= \frac{1}{L}\sum_{l=1}^{L} \Big(
    \mathcal{L}_c(y_l^c, \hat{y}_l^c)
    + \mbox{MSE}(y_l^d, \hat{y}_l^d)
    + \lambda_g \, \mbox{CE}(y_l^g, \hat{y}_l^g)
\Big), \nonumber
\end{align}
}
where $\lambda_g$ controls the contribution of the grasp loss, and $\mathcal{L}_c$ denotes the classification loss, implemented as either Cross Entropy or Focal Loss depending on the displacement window and the characteristics of the dataset.

To further improve temporal discrimination around the touch moment, we modify the supervision within the displacement window. In the standard formulation, all frames within this window are given a hard label of 1 and treated equally as ground truth. Instead, we adopt a Gaussian soft-label, where the target value decreases with distance from the annotated touch frame. This provides a smoother supervisory signal and encourages the model to concentrate its confidence near the true moment of contact rather than spreading it uniformly across neighboring frames.
During temporal offset refinement, where predicted displacement offsets are applied to adjust the temporal location of each event, we use bilinear interpolation to distribute fractional offsets across adjacent frames, preventing quantization artifacts. We further modulate the offset with a Gaussian attenuation term to ensure that large displacement values do not introduce unstable shifts in the refined score distribution. For convenience, we refer to this refinement procedure as Gauss-TOR.

\section{Experimental Analysis}
This section describes the experimental setup on TouchMoment, including implementation details, evaluation metrics, and the baseline methods used for comparison. We then present quantitative results followed by an ablation study to assess the contribution of each component in our approach.

\setlength{\tabcolsep}{4pt}
\begin{table*}[!t]
    \centering

    \begin{subtable}{\textwidth}
        \centering
        \label{tab:hoi4d_result}
        \caption{HOI4D result.}

        \begin{tabular}{@{}lllll|llll@{}}
        \toprule
                    & \multicolumn{4}{c}{Without using NMS}                       & \multicolumn{4}{c}{With NMS/SNMS}            \\ 
                    \cmidrule(lr){2-9}
                    & mAP&         $\delta$=0       & $\delta$=1    & $\delta$=2       & mAP           & $\delta$=0    & $\delta$=1    & $\delta$=2    \\
        E2E Spot & 8.71         & 6.44             & 8.95          & 10.73          & 15.37         & 2.01          & 18.10         & 26.01         \\
        UGLF     & 19.65        & \ul{13.27}       & 20.51         & 25.16          & \ul{31.85}    & \ul{6.17}     & \ul{37.75}    & \ul{51.63}    \\
        T-DEED   & \ul{20.32}   & 12.49            & \ul{21.43}    & \ul{27.04}     & 29.67         & 4.80          & 33.91         & 50.31         \\ \midrule
        \rowcolor{gray!20}
        Ours     & \tb{32.89}\gain{+12.57}    & \tb{14.25}\gain{+0.98}   & \tb{35.94}\gain{+14.51}   & \tb{48.47}\gain{+21.43}   & \tb{36.47}\gain{+4.62} & \tb{6.55}\gain{+0.38}     & \tb{42.89}\gain{+5.14}    & \tb{59.99}\gain{+8.36} \\ \bottomrule
        \end{tabular}
    \end{subtable}

    \vspace{1em}  

    \begin{subtable}{\textwidth}
        \centering
        \caption{TACO result.}
        \label{tab:taco_result}

        \begin{tabular}{@{}lllll|llll@{}}
        \toprule
                 & \multicolumn{4}{c}{Without using NMS}                                      & \multicolumn{4}{c}{With NMS/SNMS}                                 \\ 
                 \cmidrule(lr){2-9}
                 & mAP&         $\delta$=0    & $\delta$=1      & $\delta$=2       & mAP            & $\delta$=0    & $\delta$=1    & $\delta$=2  \\
        E2E Spot & 16.03        & 11.85       & 16.11           & 20.14            & 27.2           & 5.64          & 32.38         & 43.59       \\
        UGLF     & \ul{19.83}   & \ul{12.95}  & \ul{20.65}      & \ul{25.90}       & \ul{31.26}     & 6.33          & \ul{36.39}    & \ul{51.05}  \\
        T-DEED   & 17.25        & 12.56       & 15.97           & 20.60            & 27.95          & \ul{7.56}     & 33.59         & 42.69       \\ \midrule
        \rowcolor{gray!20}
        Ours     & \tb{41.08}\gain{+21.25} & \tb{24.25}\gain{+11.3} & \tb{43.47}\gain{+22.82} & \tb{55.51}\gain{+29.61} & \tb{48.38}\gain{+16.12}    & \tb{16.78}\gain{+9.22} & \tb{56.18}\gain{+19.79} & \tb{72.18}\gain{+21.13} \\ \bottomrule
        \end{tabular}
    \end{subtable}

    \caption{Quantitative comparison with end-to-end baselines on touch event detection on HOI4D and TACO. The best performing measures are highlighted in \tb{bold}, and second best measures are in \ul{underline}.}
    \label{tab.results}
\end{table*}

\subsection{Implementation Details}
For consistency with prior baselines and to accommodate hardware constraints, we train all models using clips of length $L=40$ frames, a batch size of 6, and 5000 clips per epoch for a total of 50 epochs. We use the AdamW optimizer \cite{adamw} with an initial learning rate of $4{\times}10^{-4}$, along with a three-epoch linear warm-up followed by cosine annealing. The backbone is RegNet-Y 800MF \cite{regnety}, initialized with ImageNet-pretrained weights from the \texttt{timm} library \cite{timm}. Hand patches are enlarged by a factor of 1.2 relative to the detected bounding box, padded to a square, and resized to $224\times224$. The feature dimension is set to $C=768$ with a downscaling factor of two.

Dataset-specific configurations are as follows. HOI4D contains temporally sparse touch events, so we adopt a displacement window of four and use Focal Loss \cite{focalloss} with $\alpha=0.9$ and $\gamma=2$. TACO, in contrast, includes frequent two-handed interactions where left- and right-hand touches occur in close temporal proximity. This requires a smaller displacement window of one and a weighted cross-entropy loss (weight 5.0) to reduce ambiguous supervision. The grasp loss weight is fixed to $\lambda_g=0.2$ for all experiments.

In our experiments, we found that applying MixUp \cite{mixup} significantly degraded performance, reducing mAP across thresholds by 7.37\%. We attribute this drop to the additional noise introduced in hand–object regions when two sequences are blended. Consequently, MixUp is excluded from our final model, and all remaining model settings and data augmentation strategies follow \cite{tdeed}.

\subsection{Evaluation Metrics}
We follow the standard precise event spotting protocol and report AP for the touch class at tolerance thresholds~$\delta \in \{0,1,2\}$. A prediction is counted as correct if it falls within~$\delta$ frames of the annotated touch moment. Unlike sports-based spotting benchmarks, where thresholds of \mbox{$\delta \leq 4$} are commonly used, touch events in egocentric video often occur in close temporal proximity. These low tolerance values emphasize the model's ability to localize touch events at near frame-level precision.

\subsection{Baseline Method}
We compare our method against three end-to-end baselines: E2E-Spot, T-DEED, and UGLF \cite{e2e_spot,tdeed,uglf}. Other action-spotting models such as ASTRA, COMEDIAN, and Soares \cite{astra,comedian,soares} are excluded because they are designed specifically for the SoccerNetv2 benchmark and depend on Baidu’s pre-extracted features \cite{baidu}, which are not applicable in our setting. All baselines in our evaluation are trained end-to-end from raw frames.

For fair comparison, we apply the Soft Non-Maximum Suppression (SNMS) \cite{softnms} used in T-DEED to all baseline models. We report AP at tolerance thresholds $\delta \in \{0,1,2\}$, as well as the mean AP averaged over these thresholds (mAP). For completeness, we provide results both with and without applying NMS/SNMS.

\subsection{Experiment Results}
\cref{tab.results} presents touch spotting performance on the HOI4D and TACO subsets of TouchMoment. We report results both without and with NMS/SNMS to distinguish raw temporal precision from post-processed detection quality. The No NMS setting evaluates models directly on their frame-level outputs, reflecting their intrinsic ability to localize the touch moment. In contrast, NMS/SNMS suppresses redundant detections and typically improves overall mAP by reducing false positives, though it may slightly reduce AP@$\delta$=0 when closely spaced peaks are merged or shifted, an important consideration since touch events in egocentric video often occur only a few frames apart.

Among the baselines, UGLF achieves the highest AP@$\delta$=0. This outcome is expected because UGLF leverages an external vision–language detector to identify scene objects and extract object-centric features, providing strong spatial priors that help the model attend to interaction-relevant regions. Such cues are particularly beneficial for the strict $\delta$=0 setting, where correct localization depends on capturing subtle spatial differences between adjacent frames. T-DEED, by contrast, improves substantially over E2E-Spot through its displacement heads, which encourage dense candidate predictions followed by temporal offset refinement. However, without explicit instance-level information or structured knowledge of scene entities, T-DEED cannot match UGLF in most cases where precise spatial grounding is essential. This comparison underscores the benefit of incorporating structured spatial cues in general sporting-action spotters and further motivates our hand-centric formulation for egocentric touch detection.

Across most evaluation conditions, our method achieves the strongest performance. On HOI4D, we observe substantial gains at $\delta$=1 and $\delta$=2, without NMS, our method outperforms the best baseline by 14.51 AP@$\delta$=1 and 21.43 AP@$\delta$=2, yielding a 12.57 mAP improvement. With NMS/SNMS, the gains remain pronounced, reaching 36.47 mAP and 59.99 AP@$\delta$=2, the highest among all methods.

On TACO, which involves denser and faster hand–object interactions, our method again delivers consistent improvements. Without NMS, it reaches 41.08 mAP, surpassing the best baseline by 21.25, with gains at all thresholds: 11.3 on AP@$\delta$=0, 22.82 on AP@$\delta$=1, and 29.61 on
AP@$\delta$=2. With NMS/SNMS applied, performance further increases to 48.38 mAP and 72.18 AP@$\delta$=2, exceeding the strongest baseline by 16.12 and 21.13, respectively.

\begin{figure*}[!t]
    \centering
    \begin{subfigure}{0.93\linewidth}
        \centering
        \includegraphics[width=0.96\linewidth]{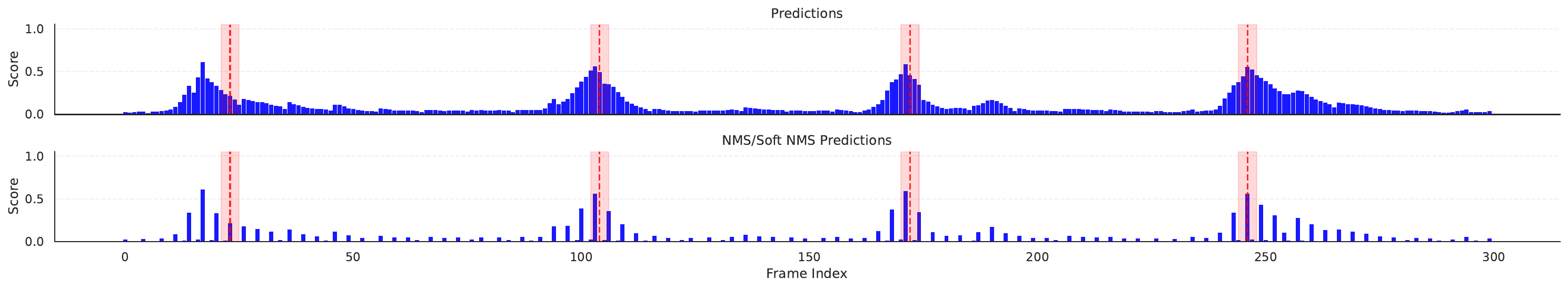}
        \caption{}
        \label{fig:hoi4d_pred}
    \end{subfigure}

    \vspace{0.5em}

    \begin{subfigure}{0.93\linewidth}
        \centering
        \includegraphics[width=0.96\linewidth]{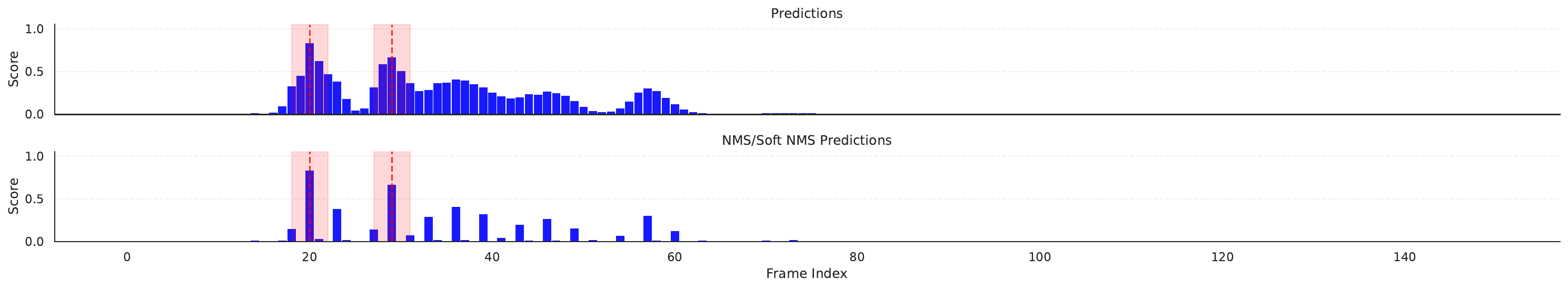}
        \caption{}
        \label{fig:taco_pred}
    \end{subfigure}

    \caption{\textbf{Qualitative examples of T-DEED with HiCE on HOI4D (a) and TACO (b)}. For each example, the top plot shows raw prediction score and the bottom plot shows predictions after NMS/SNMS. Ground-truth touch frames are indicated by red dashed lines, with a tolerance window of $\delta=2$ frames.
    }
    \label{fig.qual_res}
\end{figure*}

\subsection{Ablation Studies and Qualitative Results}
\cref{tab.ablation} summarizes ablations on HOI4D and TACO, evaluating the effect of individual model components, clip length, and context size. Using HiCE alone provides a strong baseline, but the full model yields the best performance across both datasets. On HOI4D, removing soft labels results in a clear performance drop. HOI4D is trained with a large displacement window (size 4), meaning many frames surrounding the true touch moment are annotated as positives. Without soft labels, these near-touch frames become indistinguishable from the true contact frame. Soft labels help resolve this ambiguity by assigning smoothly decaying supervision around the ground truth, enabling the model to better isolate the precise moment of contact. HOI4D also benefits considerably from grasp supervision: the dataset is larger, more diverse, and contains richer hand–object configurations, allowing the grasp branch to regularize the hand-centric representation rather than overfit.
On TACO, however, the contributions of each constituent component are different.  TACO is only about one quarter of the size of HOI4D and uses a much smaller displacement window (Size 1), resulting in far fewer positive frames and less variation. In this more limited setting, the grasp branch tends to distribute predictions across neighboring frames to model the grasp pattern, which improves robustness across thresholds but slightly dilutes exact-frame precision at $\delta=0$. Soft labels also provide less benefit on TACO because near-touch frames are tightly clustered and the narrow displacement window already enforces sharp supervision. Temporal offset refinement (Gauss-TOR) consistently improves accuracy on both datasets, confirming its role in stabilizing offset predictions and producing sharper detection peaks. For clip length and context size, $L=40$ and $\times$1.2 yield the best overall balance, as insufficient context harms detection while excessive context dilutes the relevant signal.
\begin{table}[h]
\centering
\footnotesize
\setlength{\tabcolsep}{6pt}
\renewcommand{\arraystretch}{1.0}
\begin{tabular}{@{}lrrrr@{}}
\toprule
                          & \multicolumn{2}{c}{\tb{HOI4D}} & \multicolumn{2}{c}{\tb{TACO}} \\
                          & mAP    & $\delta$=0  & mAP    & $\delta$=0  \\
\midrule
\multicolumn{5}{l}{\textit{\quad Model components}} \\
Proposed               & \tb{32.89} & \tb{14.25} & 41.08 & \tb{24.25} \\
w/o Grasp Loss            & 26.96       & 11.70            & \tb{44.13}       & 21.87            \\
w/o Gauss-TOR             & 30.3        & 14.23            & 33.09       & 22.74            \\
w/o Soft Label            & 23.38       & 11.66            & 41.09       & 18.84            \\
only HiCE                 & 21.10       & 11.37            & 30.79       & 15.97            \\
\midrule
\multicolumn{5}{l}{\textit{\quad Clip length}} \\
$L=25$                      & 17.16  & 7.99        & \tb{44.96}  & 21.70       \\
$L=40$ (proposed)               & 32.89  & \tb{14.25}       & 41.08  & \tb{24.25}       \\
$L=50$                      & 33.80  & 13.20       & 40.54  & 16.33       \\
$L=80$                      & \tb{34.93}  & 12.13       & 37.63  & 19.16       \\
\midrule
\multicolumn{5}{l}{\textit{\quad Hand context size}} \\
$\times$1.0               & 31.40  & 13.94       & 32.15  & 14.95       \\
$\times$1.2 (proposed)        & \tb{32.89}  & \tb{14.25}       & \tb{41.08}  & \tb{24.25}       \\
$\times$1.5               & 32.22  & 12.81       & 33.41  & 19.57       \\
\bottomrule
\end{tabular}
\caption{\tb{Ablation study for TouchMoment}. We evaluate the effect of individual model components, clip length $L$, and context size on HOI4D and TACO subsets. Bold indicates best results.}
\label{tab.ablation}
\end{table}

\cref{fig.qual_res} presents qualitative results for T-DEED with HiCE on HOI4D (top) and TACO (bottom). Each example contains two plots: the upper plot displays the raw prediction scores, while the lower plot shows the predictions after NMS/SNMS. In each plot, blue bars represent the predicted touch scores (ranging from 0 to 1). The red dashed line marks the ground-truth touch frame, and the shaded red region denotes the tolerance window of $\delta=2$ frames. Predictions that fall within this region are counted as true positives. In the HOI4D example, training with a displacement window of 4 generates multiple nearby candidate peaks that consolidate tightly around the ground-truth after applying displacement offset and post-processing. In the TACO example, despite two touch events occurring only nine frames apart (300 ms), the model produces two distinct peaks, demonstrating HiCE's ability to resolve closely-spaced temporal events.

\section{Conclusions and Future Work}
In this paper, we tackled the task of detecting the precise moment when a hand makes contact with an object in egocentric video. We introduced HiCE, a hand-informed context enhancement module that augments frame-level features with specialized spatiotemporal hand representations. This approach leverages cross-attention to fuse hand-centric cues with global scene context, enabling models to discriminate true touch moments from visually similar near-contact frames. Alongside the architectural innovation, we presented TouchMoment, a dataset comprising 4,021 videos with 8,456 annotated touch moments across diverse manipulation scenarios. Our experiments on TouchMoment and existing egocentric datasets demonstrate HiCE's effectiveness in achieving frame-accurate touch detection under strict temporal tolerance.

{\small    \myheading{Acknowledgements:} This work was funded by the Australian Institute for Machine Learning (Adelaide University) and the Centre for Augmented Reasoning, an initiative by the Department of Education, Australian Government.

{
    \small
    \bibliographystyle{ieeenat_fullname}
    \bibliography{longstrings, references}
}


\end{document}